\documentclass{vldb}

\usepackage{graphicx}
\usepackage{xcolor}
\usepackage{url}
\usepackage{balance} 
\usepackage{amsfonts,amssymb}
\usepackage{algpseudocode}
\usepackage{algorithm}
\usepackage{algorithmicx}
\usepackage{amsmath}
\usepackage[makeroom]{cancel}
\usepackage{caption}
\usepackage{hyperref}
\usepackage{listings}
\usepackage{paralist}
\usepackage{enumitem}
\usepackage{pgfplots}
\pgfplotsset{compat=1.14}
\usepackage{siunitx}
\usepackage{verbatim}
\usepackage{nccmath}

\usepackage{booktabs} 
\usepackage{placeins}

\usepackage{array}
\usepackage{arydshln}
\makeatother
\vldbTitle{DJEnsemble: On the Selection of a Disjoint Ensemble of Black-box Deep Learning Spatio-temporal Models}
\vldbAuthors{Yania Molina Souto, Rafael Silva Pereira, Roc\'io Zorrilla, Anderson Chaves, Brian Tsan, Florin Rusu, Eduardo Ogasawara, Artur Ziviani, Fabio Porto }
\vldbDOI{https://doi.org/10.14778/xxxxxxx.xxxxxxx}
\vldbVolume{12}
\vldbNumber{xxx}
\vldbYear{2021}

\mathchardef\mhyphen="2D

\algnewcommand\algorithmicforeach{\textbf{for each:}}
\algnewcommand\ForEach{\item[ \algorithmicforeach]}

\begin{document}


\title{DJEnsemble: On the Selection of a Disjoint Ensemble of Deep Learning Black-Box Spatio-temporal Models}



%
%
%
%

\numberofauthors{1} 
\author{
%
%
\alignauthor 
Yania Molina Souto$^{1}$,
Rafael Pereira$^{1}$,
Roc\'io Zorrilla$^{1}$,
Anderson Chaves$^{1}$,
Brian Tsan$^{2}$,
Florin Rusu$^{2}$,
Eduardo Ogasawara$^{3}$,
Artur Ziviani$^{1}$,
Fabio Porto$^{1}$\\
\affaddr{$^{1}$LNCC-DEXL, Petropolis, Brazil}\hspace*{0.25cm}
\affaddr{$^{2}$UC Merced, Merced, CA, USA}\hspace*{0.25cm}
\affaddr{$^{3}$CEFET-RJ, Rio de Janeiro, Brazil}\\
\email{\{yaniams, rpereira, romizc, achaves\}@lncc.br, \{btsan, frusu\}@ucmerced.edu, eogasawara@ieee.org, \{ziviani, fporto\}@lncc.br}
}

\date{1 April 2020}


\maketitle

\begin{abstract}
Consider a set of black-box models -- each of them independently trained on a different dataset -- answering the same predictive spatio-temporal query. Being built in isolation, each model traverses its own life-cycle until it is deployed to production. As such, these competitive models learn data patterns from different datasets and face independent hyper-parameter tuning. In order to answer the query, the set of black-box predictors has to be ensembled and allocated to the spatio-temporal query region. However, computing an optimal ensemble is a complex task that involves selecting the appropriate models and defining an effective allocation function that maps the models to the query region.

In this paper, we present a cost-based approach for the automatic selection and allocation of a disjoint ensemble of black-box predictors to answer predictive spatio-temporal queries. Our approach is divided into two stages---offline and online. During the offline stage, we preprocess the predictive domain data -- transforming it into an aligned non-regular grid -- and the black-box models---computing their spatio-temporal learning function. In the online stage, we compute and execute a DJEnsemble plan, which minimizes a multivariate cost function based on estimates for the prediction error and the execution cost---producing a model spatial allocation matrix. We conduct a set of extensive experiments that evaluate the DJEnsemble approach and highlight its efficiency. We show that our cost model produces plans that are close to the best plan. When compared against the traditional ensemble approach, DJEnsemble achieves up to $4X$ improvement in execution time and almost $9X$ improvement in prediction accuracy.
\end{abstract}

\section{Introduction}
\label{sec:Introduction}
As AI expands into wide economical and societal activities, an increasing number of AI models are developed and embedded in diverse applications, ranging from finance \cite{Zheng:2019} to medical patient diagnosis \cite{LiuXiaonxuan:2019}. A particular class of AI models aims to predict spatio-temporal phenomena, such as weather forecast and urban transportation \cite{Cheng:2018}. Examples include AccuWeather, Inc. \cite{Accuweather2018,sciencedaily} and Zhang et al. \cite{Zhang2017} for temperature prediction; Google AI \cite{googleblogAi} and Souto et al. \cite{Yania:2018} for rainfall forecast; and models to alert against traffic accidents that combine environmental attributes, road conditions, and satellite images \cite{Zhuoning:2018,Najjar:2017}. In these scenarios, a learner captures data patterns that are both spatially and temporally correlated to the prediction. Moreover, the spatio-temporal learner improves its prediction accuracy by learning data patterns from observations captured at neighboring locations. These characteristics add special challenges to the traditional prediction problem.

In our research experience, we have noticed that operational data are shared within large organizations according to the business needs and following proper access policies. That is not the case with models whose sharing is less common. Data scientists tend to build and use their models for local applications, unintentionally ``hiding'' them from a wider use within the organization. To address this issue, we consider a paradigm where both data and models are shared in production. The assumption is that models have been trained and validated independently -- as in a traditional machine learning life-cycle \cite{Zaharia:2018} -- and are integrated into applications as black-box functions. In this paradigm, challenges emerge in the model selection process to answer a given prediction query. Black-box models exhibit different performance due to variation in the predictor's architecture, hyper-parameters' configuration, and the data samples observed during training \cite{Kang:2019}. Even when the hyper-parameters have been diligently tuned, newly arriving data may reflect new patterns---which can flag models for updating. Nevertheless, as it has been argued by Leszczynski et al. \cite{Leszczynski:2020}, it is common to keep the existing predictor in order to avoid instability in production. Hence, when sharing models in production, one must account for their varying performance in different regions of the domain.

To illustrate this paradigm, consider a weather forecast scenario in Brazil, in which three spatio-temporal predictors (STP) are available. The first STP -- by Souto et al. \cite{Yania:2018} -- is a ConvLSTM model for temperature and rainfall prediction trained on a slice of the CFSR dataset \cite{cfsr} covering a region of the Brazilian territory. The second STP is STCONVS2S \cite{nascimento2019stconvs2s}, a deep learning model for spatio-temporal prediction with a different learning approach, trained on the same dataset. Finally, the third STP \cite{Oliveira:2015} implements an artificial neural network trained with data produced by operational numerical weather prediction models covering the city of Rio de Janeiro. When a vegetable farmer from a mountainous region at an altitude of 800m in the Rio de Janeiro state looks for the weather forecast for her farm, how can she take advantage of the availability of these different predictors to obtain the most accurate prediction? While the first two STPs cover the farm location, the corresponding data are combined with data from many other locations that are far away. The third STP includes only data that are spatially closer to the farm. However, the difference in altitude may adversely impact its accuracy. The temporal aspect is also important because the data used for the third STP has a higher frequency. Lastly, although the first two STPs are built on the same training data, their prediction time and memory usage are highly different.

\begin{figure}[!ht]
	\centering
	\includegraphics[scale=0.4]{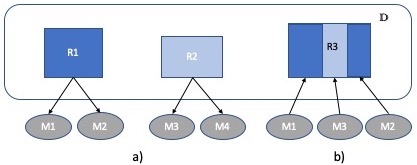} 
	\caption{In a), STPs $M_1$ and $M_2$ are trained on region $R_1$, while STPs $M_3$ and $M_4$ are trained on region $R_2$. In b), the optimal allocation of models $M_1$, $M_3$, and $M_2$ to answer a query over a different region $R_3$ is depicted.}
	\label{fig:Model-Selection}
\end{figure}

Abstractly, assume that a user has access to multiple black-box STP models that can answer predictive spatio-temporal queries. Then, the question we want to answer is: ``How to select the STP -- or STP combination -- that gives the optimal performance, where performance is measured as a function of multiple parameters, including accuracy, execution time, and resource utilization?'' Our approach is to formulate and solve an optimization problem that finds the optimal STP ensemble of black-box predictors that minimizes a multivariate cost function. The solution specifies an allocation of the selected predictor's spatial frames to the query region that forms a \textbf{D}is\textbf{J}oint \textbf{Ensemble} (DJEnsemble) allocation, as depicted in Figure \ref{fig:Model-Selection}. However, identifying such an ensemble is a hard, problem as it involves: (i) estimating the prediction accuracy of each black-box predictor at the query region; (ii) defining a black-box model ensembling strategy; (iii) finding the ensemble plan that minimizes the cost function; and (iv) planning for the execution.

In this paper, we present DJEnsemble, a novel method to solve the STP query optimization problem in order to build an ensemble that maximizes the performance of answering predictive spatio-temporal queries. Our method has two phases---offline and online. During the offline phase, we perform the following steps for each spatio-temporal domain:
\begin{itemize}[leftmargin=*,noitemsep,nolistsep]
 \item Cluster the domain time-series (Section \ref{sec:formal}) using a feature based-approach \cite{Aghabozorgi:2015} in order to identify regions with similar data distribution. 
 \item Partition the spatio-temporal domain into a disjoint grid of time-series tiles sharing the same cluster and having similar data distribution.
 \item Compute a centroid time-series as a representative for the set of time-series in every tile, to be used in data distribution distance computation.
 \item For every black-box model, compute a learning curve that predicts its behavior as a function of the distance between the data distribution in the training and query regions, in order to predict the model performance on unseen data.
\end{itemize}

Once the offline phase is complete, we can answer spatio-temporal predictive queries in the online phase as follows:
\begin{itemize}[leftmargin=*,noitemsep,nolistsep]
 \item Find the tiles whose spatio-temporal frame intersects with the query region.
 \item Select candidate models for every tile.
 \item Allocate each tile to the model that minimizes a multivariate cost function that includes the estimated error and the execution metrics.
 \item Build a disjoint STP ensemble plan based on the computed allocations.
 \item Execute the ensemble according to the plan and compose the overall prediction.
\end{itemize}

We perform extensive experiments that evaluate our approach on two real and one synthetic datasets, and six queries over a set of 36 STPs. Firstly, we show that the learning curve correctly approximates the prediction error as a function of data distribution distances. Next, we show that the optimization procedure implemented by DJEnsemble is capable of selecting a good ensemble plan out of a large number of possible predictor allocations. The approach is resilient to different scenarios involving the predictors' architecture and training datasets. Finally, we compare the results obtained by DJEnsemble against 5 other models, including both a traditional and a stacking ensemble. DJEnsemble achieves an accuracy improvement of up to $9X$ and it reduces query prediction time by a factor of up to $4X$.

We can summarize our contributions as follows:
\begin{compactitem}
    \item A \textit{data distribution based methodology} to select spatio-temporal models for building ensembles
    \item A \textit{multi-variate cost model and allocation approach} to support the DJEnsemble methodology
    \item A \textit{comprehensive experimental evaluation} over two real datasets of meteorological observations and a synthetic dataset, 6 queries, and 36 models 
\end{compactitem}


\section {Problem Formulation}
\label{sec:formal}

Consider a spatial domain $\mathbb{D}(D,V)$, where $D=\{p_1, p_2, \allowbreak \ldots, p_n\}$ is its discretization into a set of localized 2D-points $p_i(x_i,y_i)$, with $x_i$ and $y_i$ being spatial coordinates. At every point $p_i \in D$, observations are recorded as a time series $V$. Thus, an observation in $V$ for spatial point $p_i$ and time $t_j$ is referred to as $v_{i,j}$. Additionally, we assume there are multiple black-box spatio-temporal predictors (STP) available. The STPs are trained on datasets structured as a set of spatio-temporal series $\mathbb{ST}=(\langle x,y \rangle, V)$, where $\langle x,y\rangle$ corresponds to the spatial dimensions and $V$ is a time-series of observations \cite{Campisano:2018}. For STPs training and prediction, the data are preprocessed as a list of bi-dimensional matrices (i.e., frames) that hold data corresponding to a time instant. Every cell of the matrix holds a time-series at the corresponding spatio-temporal location. An STP receives a sequence of input frames $I= \langle I_1, I_2, \ldots ,I_n\rangle$ and produces a list of predictions as $K$ output frames $O=\langle O_{n+1},\ldots,O_{n+k}\rangle $. Data in $I$ contain the spatio-temporal observations from which predictions are made. For example, $I$ may refer to the temperature in a spatial region during the last five days (i.e., number of $I$ frames), and predicts the temperature in the same region for the next three days (i.e., three $O$ frames).

A user issues a spatio-temporal query $Q$ defined as:
\begin{equation}
Q=\{R, ptime, V_q, Input, M_e\}
\end{equation}
where $R$ is a 2D spatial region defined over the same domain $\mathbb{D}$, $R \subseteq D$. A region $R=(start, height, width)$ specifies a 2D location, $start$, and its orthogonal extension in a coordinate system. The size of $R$ is given by its area, $height \times width$. $R$ can be split into tiles $R_\#$, $1 \le \# \le n$, such that $\bigcup_{\#=1}^{n} R_\# = R$ and $R_i \cap R_j = \emptyset, 1 \leq i,j \leq n, i \neq j$. $ptime$ defines the number of time-steps to be predicted, while $V_q$ is the quantity to be predicted by the query, whose values are drawn from the time-series $V$. $Input$ is a dataset of time-series $V$ in the spatial region delimited by $R$ and structured as frames $I$, which are input to a predictor. Finally, $M_e$ is a performance metric, such as \emph{root mean square error}, \emph{prediction execution elapsed time}, etc. In the query corresponding to the farm scenario, $R$ specifies a mountainous area outside the city of Rio de Janeiro, $ptime$ is three days, $V_q$ is the time-series of temperatures, $Input$ is a dataset of temperatures in $R$, and $M_e$ is the root mean square error function.

To compute the predictions $V_q$, we have a set of STPs $M=\{m_1, m_2, \ldots m_s\}$. Every model $m_i \in M$ is associated with metadata and its training dataset (e.g., a slice of the CFSR dataset). Thus, a model is specified as:
\begin{equation}
m(Id,dataset,region,error\mhyphen function,\\ frame \mhyphen size)
\end{equation}
where $region$ identifies a 3D \textit{spatio-temporal} training region, $error \mhyphen function$ specifies a learning curve as the error estimate for a given data distribution distance (Section \ref{sec:error-function}), and $frame\mhyphen size$ is the predictor's frame area.

We formalize the \emph{optimization of spatio-temporal predictive queries} (OSTEMPQ) problem as follows. Given a spatial domain $\mathbb{D}$ and its discretization $D$, a spatio-temporal query $Q$, and a set of STPs $M$, determine the optimal allocation $A=(R_\#,M')$ of STPs to the query tiles, where $M' \subseteq M$ and $\bigcup_{\#=1}^{n} R_\# = R$. The execution of the models $M'$ from allocation $A$ produces a spatio-temporal prediction $C$ that satisfies the constraints in $Q$ and has an execution cost that minimizes $Me$. We solve the OSTEMPQ problem under the following constraints:
\begin{align}
	\label{eq:solutionConstraint}
	(i) & \forall \,\, R_i \in R_\#, \ \exists \ \textrm{model} \,\, m_j\ \in\ M' \textrm{such that}\ A(R_i, m_j) \nonumber \\ 
	(ii) & A(R_j, m_i) \wedge A(R_j, m_k)\,\, \textrm{if only if} \,\, i=k
	\end{align}
where the allocation $A$ is a non-injective non-surjective function.
Observe that, although we allocate only one model per tile (ii), a query can cover multiple tiles, leading to an allocation comprising a disjoint ensemble of models.

\section{Preliminaries}\label{sec:preliminaries}
In this section, we present the spatio-temporal predictors considered in this work and the Generalized Lambda Distribution (GLD) probability density function (pdf) used to cluster data.

\textbf{Deep learning models for spatio-temporal predictions.}
Deep learning spatio-temporal models have been extensively used in video and image analysis. Conv3D \cite{Tran_2015_ICCV} is the first successful convolution architecture to process large video datasets using only 3D convolution layers. In follow-up work by Tran et al. \cite{Tran:2018}, the authors suggest factorizing the space and temporal filters in two separate and consecutive (2+1)D components. More inline with our work is the Convolution LSTM (ConvLSTM) architecture proposed by Shi et al. \cite{shi2015convolutional}. This architecture learns the spatial signals using convolution operators, which are followed by a recurrent neural network layer applied to the prediction. The Conv(2+1)D and ConvLSTM approaches are combined in the more recent STConvS2S model \cite{nascimento2019stconvs2s}. 
 
We consider spatio-temporal deep learning models to solve regression problems, where the input is a list of fixed-size frames and the output is also a list of frames of the same size as the input. The number of input frames determines the rate of the input temporal signal. In this scenario, the above classification models are not a direct fit. A possible solution is to apply the ARIMA (AutoRegressive Integrated Moving Average) model \cite{Box:2015} to evaluate an autoregressive prediction for each time-series in the query region. In this paper, we adopt the ConvLSTM architecture as our baseline. Given a particular ConvLSTM model with a fixed-size frame and a spatial area where predictions are computed, multiple invocations of the model may be necessary to cover the entire area. 

\textbf{Modeling data distributions with GLDs.}
During training, a learner captures the data patterns in the input dataset \cite{Shalev2014}. Different approaches to learn data distributions \cite{Cohen-Shapira2019} and extract meta-features from the training data \cite{Vainshtein2018AHA,Aghabozorgi:2015} are proposed in the literature. We adopt the Generalized Lambda Distribution (GLD) probability density function (pdf) \cite{Ramberg:1974} because it can model an entire family of data distributions, such as Gaussian, Logarithm, and Exponential \cite{Chalabi:2012}. GLD encodes different data distributions through the specification of the lambda parameter representing statistical moments, where $\lambda_{1}$ and $\lambda_{2}$ determine the location (i.e., mean) and scale (i.e., standard deviation) parameters, while $\lambda_{3}$ and $\lambda_{4}$ determine the skew and kurtosis of the distribution, respectively. Then, a GLD is represented as $GLD(\lambda_{1},\lambda_{2},\lambda_{3},\lambda_{4})$, known as the \textit{RS} parametrization \cite{Ramberg:1974}. We fit GLDs to the distribution in the time-series at each spatial position and time seasonality interval \cite{liu2020}. Then, we use the $\lambda$ parameters to identify regions sharing similar distributions (Section \ref{sec:clustering}).

\begin{figure}[!ht]
	\centering
	\includegraphics[width=0.45\textwidth]{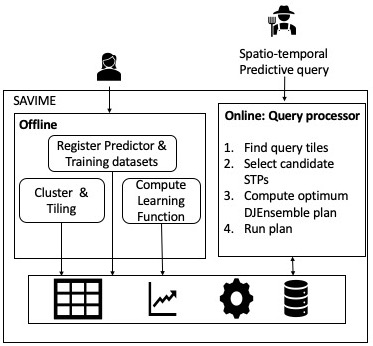}
 \caption{DJEnsemble architecture in SAVIME.}
 \label{fig:savimeextended}
\end{figure}

\section{The DJEnsemble Approach}

We propose the DJEnsemble approach to solve the OSTEMPQ problem. To the best of our knowledge, DJEnsemble is the first work to solve the optimal allocation of black-box models to answer predictive spatio-temporal queries. DJEnsemble has an offline and an online phase. The goal of the offline preprocessing phase is to prepare the data and the models for effective predictive querying. This is a common strategy to reduce query execution time and improve prediction accuracy. While time-consuming for certain datasets and models, the offline stage is executed only once. In the online phase, a cost function guides the search for ensembles that answer a given query optimally. Figure \ref{fig:savimeextended} depicts the high-level architecture of the DJEmsemble, which is implemented as an extension of the SAVIME multidimensional array processing \cite{lustosa:2019}.

\subsection{Offline: Preprocessing}\label{sec:offline}

In the offline phase, we preprocess data for a given domain, cluster regions with similar distributions, and repartition the domain into aligned non-regular tiles. A representative time-series that models the data distribution is associated with every tile. Additionally, we build a learning function to estimate a model's performance on different areas of the domain for every model in the set $M$.

\subsubsection{Clustering}\label{sec:clustering}

We perform clustering in order to group time-series with similar data distribution, adopting a feature-based approach \cite{Aghabozorgi:2015}. The GLD function is used as a mechanism to compute the time-series features---exposed through its lambda parameters. We fit a GLD function to every time-series $V_i$ from domain $D$ and associate the four $\lambda$ parameters (i.e., time-series features) to it. If $V_i$ exhibits seasonality, we fit a separate GLD function to each season. Thus, domain $D$ is represented as $D_t (p ,V, \lambda_1,\lambda_2, \lambda_3, \lambda_4)$. After determining $D_t$, we cluster the time-series that have similar $\lambda$ values together, using a \emph{k-means} algorithm. We use \emph{silhouette analysis} to guide the choice of $k$. Therefore, dataset $D_t$ is transformed into $D_c(p ,V, \lambda_1,\lambda_2, \lambda_3, \lambda_4, cid)$, where $cid$ identifies the cluster each point belongs.

\begin{figure}[!ht]
	\centering
	\includegraphics[width=0.35\textwidth]{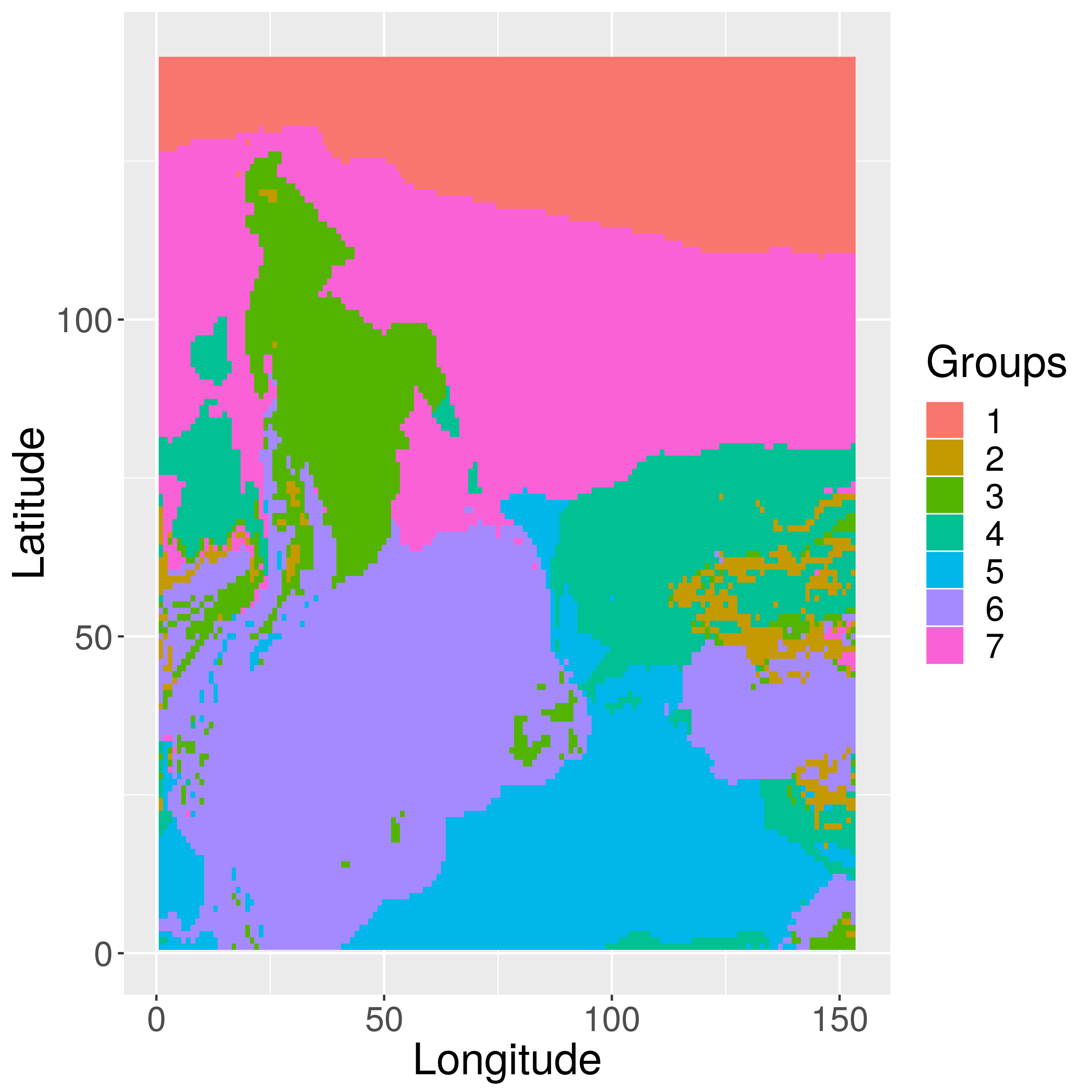} 
	\caption{The clusters of a spatial domain containing temperature readings. Silhouette analysis indicates 7 clusters.}
	\label{fig:cluster}
\end{figure}

Figure \ref{fig:cluster} depicts the results of applying k-means clustering to a spatio-temporal dataset of temperature readings. Notice the non-convex regions with a mixed distribution defined by the clusters. Achieving good estimation accuracy by regular frame STPs on this sort of region is challenging.

\subsubsection{Tiling}\label{sec:tiling}

Tiling has been introduced in the context of multidimensional array database systems. In \cite{Furtado:1999}, different tiling approaches are discussed. Among them, non-aligned tiling divides a multidimensional array into disjoint tiles, where the vertices of a tile do not intersect with those of neighboring tiles. In DJEnsemble, we adopt this tiling strategy in order to partition the domain into tiles of time-series sharing the same cluster id and, as a consequence, exhibiting time-series with similar data distribution. The tiling algorithm is inspired by the object detection method in YOLO \cite{Yolo:2016}. We start from an arbitrary point $p_i(\langle x_i, y_i\rangle, V_i)$ and aggregate neighboring points to form a hypercube---as long as they belong to the same cluster as $p_i$. We repeat this process until every point in $D$ has formed a tile \textit{Tile (id,}$\langle$\textit{coordinates}$\rangle$, \textit{centroid)}. $coordinates$ corresponds to the 3D spatial region covered by the tile, while $centroid$ is the closest time-series to all the other series in the tile. We use $centroid$ as a representative of a tile's data distribution in order to simplify the computation of the distance between two tiles.

\subsubsection{Estimating model performance on query region} \label{sec:error-function}

In addition to preprocessing the domain data, we estimate the performance of every candidate model on every region of the domain. For this, we define a data distribution-based prediction error function as follows:
\begin{equation}\label{eq:error-function}
E_{gnr}=F_\epsilon(dist_{ij}+e_i)
\end{equation}
$F_{\epsilon}(dist_{i,j}+e_i)$ is a monotonic non-decreasing function for every candidate model. $dist_{i,j}$ is the distance between the distribution in the model's training data $d_i$ and a prediction region $d_j$. $e_i$ denotes the model's generalization error, obtained on the testing dataset the model is evaluated on. We assume that the training and the testing datasets have similar distributions.

For a given model, we compute its corresponding learning function $F_{\epsilon}(dist_{i,j}+e_i)$ by fitting a polynomial model to a series of pairs $(dist,error)$. It has been shown that learning curves that estimate a model loss as a function of an increasing training dataset follow a power-law \cite{Domhan:2015}. In DJEnsemble, we consider the increasing distance from a data distribution to be a more precise measure than the size of the training dataset for predicting the estimation accuracy. Thus, our goal is to find a polynomial function that maps the data distribution distance to prediction error. The fitting process works as follows:
\begin{itemize}[leftmargin=*,noitemsep,nolistsep]
 \item Select regions with different data distributions than the one of the training dataset.
 \item Generate a modified version of these regions' data sequentially by adding Gaussian noise, $r_i=r_{i-1} + \mathcal{N}(0,\sigma), 0 \leq i \leq n-1$, with increasing $\sigma$ values.
 \item Calculate the distance between the training dataset and the modified dataset using the DTW function applied over the centroid time-series.
 \item Compute the model prediction error on the modified data.
 \item Train a polynomial regression model based on the pairs $(dist,error)$ using the fitting function $error=F_{\epsilon}(dist_{i,j}+ e_i)$. The regression model is trained in increasing order until no improvement is observed.
 \item Approximate the error of the model on unseen data by applying the fitted polynomial $F_{\epsilon}(dist_{i,j}+e_i)$ to that data.
\end{itemize}

At the end of the offline preprocessing phase, the spatio-temporal series are partitioned into tiles of homogeneous data distributions---represented by their centroid spatio-time series. Moreover, a generalization error function predictor is derived for every registered spatio-temporal prediction model---parameterized by a data distribution distance between the time-series. The latter is the fundamental mechanism in predicting the accuracy of a model on an unseen query region. It is important to observe that the learning function approach is applicable to any model type.

\subsection{Online: Query Processing}\label{sec:online}

In the online phase of the DJEnsemble, the spatio-temporal predictive query $Q$ is evaluated. The domain is partitioned into tiles $T$ and there is a set of candidate models $M$ to predict the query variable $V$. The online phase of the DJEnsemble is split into three stages---planning, execution, and post-processing. In the planning stage, a set of black-box candidate predictors are selected and the allocation matrix is computed. In the execution stage, the selected models are evaluated according to the planned allocations. Lastly, post-processing actions are taken only when necessary. The following sections discuss three solutions corresponding to the three execution stages in DJEnsemble.

\subsubsection{STP ensemble model}\label{sec:composition-approach}

\begin{figure}[!ht]
	\centering
	\includegraphics[scale=0.4]{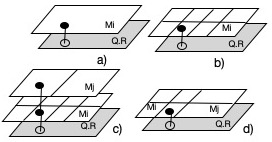} 
	\caption{Black-box spatio-temporal prediction (STP) models with a different frame size to answer a query $Q$ over region $R$: a) Single STP model, b) Single STP model with the same spatial dimensions as the query, c) Ensemble STP model, and d) DJEnsemble STP model. }
	\label{fig:st-ensemble}
\end{figure}

The single STP model approach to solve the OSTEMPQ problem is depicted in Figure \ref{fig:st-ensemble} a) and b). In a), the model $M_i$ covers completely the query region $Q.R$. Thus, a single instance of the model is sufficient to evaluate the query. In b), the area of the query region is larger than the model frame. Multiple instances of the STP model that cover the entire query region are required in this case.

The \textit{traditional ensemble} STP model \cite{Zhang:2012} -- depicted in Figure \ref{fig:st-ensemble} c) -- evaluates query $Q$ as follows. In the planning stage, it selects a subset $M'\subseteq M$ of models with testing accuracy higher than some threshold $\delta$. An allocation matrix consisting of multiple instances is built for every model in $M'$, such that the entire query region $Q.R$ is covered by every model. In the execution stage, the allocations are submitted to a prediction execution engine (e.g., TensorFlow \cite{tfxServer}). Finally, in the post-processing stage, an aggregation operation computes a linear combination of the results.

The DJEnsemble (DisJoint Ensemble) approach extends the planning phase of the traditional ensemble. In addition to identifying the set of models $M'$ that complies with the threshold $\delta$, it also computes an allocation matrix $A(R_\#,M')$ according to the constraints in Eq. \ref{eq:solutionConstraint} and the cost function \ref{eq:cost} (Figure \ref{fig:st-ensemble} d). The latter allocates the model with the minimum cost to every query subregion $R_\#$. The allocations in $A$ are executed by an execution engine and the generated predictions are composed in the query result.

In the single STP approach, the model with the highest testing accuracy is allocated to cover the entire query region $Q.R$. In order to compute the predicted value $v_{i,j+1} \in V$ for a point $p_i$ in the query region $Q.R$ at time step $t_{j+1}$, the traditional STP ensemble executes all the ensembled models and aggregates their predictions at every spatio-temporal coordinate $\langle p_i, t_{j+1}\rangle$. In the DJEnsemble STP model, every predicted value $v_{i,j+1} \in V$ in $Q.R$ is the result of a single predictor allocated over the volume containing point $p_i$. While we select the DJEnsemble model as our solution, we compare it against the other alternatives in Section \ref{sec:baseline}.
 

\subsubsection{Model composition search}

The main challenge of the DJEnsemble STP model is to compute the assignment $S=(M,A)$ of models $M$ to query subregions $A$. The computation of $A$ requires considering the allocations of models in $M$ over the query region $Q.R$. For every candidate model $m \in M$ with frame size $m_{fs}$ ($m.frame\mhyphen size$) -- a fraction of the query frame size $q_{fs}=(R.size)$ -- we can align its frame's top-left corner with any of the $p_i$ spatial positions in $Q.R$. We can repeatedly apply this procedure until all the points in $Q.R$ are considered for prediction by a model in $M$. However, this exhaustive procedure hinders the ability to execute prediction queries efficiently due to the high overhead it incurs. Instead, we partition the query domain into \textit{tiles}, as described in Section \ref{sec:tiling}. Given that each tile covers a region with time-series having similar data distribution, we pick the model whose training data distribution resembles that of the tile's centroid the most. This procedure reduces the search space for selecting candidate models to every query tile, which are considerably fewer than the number of observation points.

Given a suggested allocation, the implication of a possible difference between $m.frame\mhyphen size$ and the tile size is managed as follows. First, models are placed with the top-left corner matching that of the tile. Then, for models whose frame size is a fraction of the tile size, we place as many non-overlapping instances of the model so that the tile region is covered. Conversely, in case the model frame extends beyond the tile area, we only consider predictions on spatial points falling within the tile area.

\subsubsection{Cost function}\label{sec:cost_based}

We design a cost function to optimize model allocation as the linear combination of a model's estimated generalization error and its estimated prediction execution time. The error estimate is computed by the learning function, as described in Section \ref{sec:error-function}. The estimate for the prediction execution time is obtained by averaging the model's previously recorded execution times, leading to a unitary cost $(uc)$. Moreover, depending on the ratio between the model frame size and a tile's 2D size, a number $\lceil r \geq 1 \rceil$ of invocations of the model are required to cover all the points in the tile region. In this case, every candidate allocation $A(t_i,m_j)$ for tile $t_i$, model $m_j$, and a weighting parameter $\mu_e$, is assigned a cost given by the formula:
\begin{equation}\label{eq:cost}
 Cost_{i,j} = (1 -\mu_e) \times F_\epsilon(dist_{i,j}+\epsilon_{i}) + \mu_{e} \times \lceil r_{i,j} \rceil \times uc
\end{equation}
This cost formula normalizes the generalization error and the prediction time to $[0,1]$ intervals by dividing each value by the maximum value in the set of models, once outliers are eliminated from the set. The maximum values can be computed in a single pass over the predictions.

\subsubsection{DJEnsemble algorithm}

Algorithm \ref{alg:online} presents the DJEnsemble algorithm. It takes as input the query $Q$, the set of tiles $T$, the set of candidate models $M$, and a weight parameter $\mu_e$. The \textit{DJEnsemble} function returns the set of mappings $A(T_i,M_j)$ that satisfy the constraints in Eq. \ref{eq:solutionConstraint} and minimize the cost function detailed in Section \ref{sec:cost_based}.


\begin{algorithm}[h!]
\caption{DJEnsemble Algorithm}\label{alg:online}
\begin{algorithmic}[1] 
\Function{DJEnsemble} {$Q,T,M,\mu_e $}

\State {$query_T \gets queryTiles(Q,T)$}
\For{$q_t\ \in\ query_T \ in\ parallel$}

 /* Extract the query tile centroid /*
 \State {$q_c\ \leftarrow\ q_t.getcentroid()$}
 
 /* Min priority queue /*
 \State {$pq_i \gets \bot $}
 
 /* Compute generalization error /*
 \State {$M.estimate\_generalization\_error(M,q_t) $}
 
 /* Remove models with estimated error outliers /*
 \State {$M' \gets drop\_outlier\_model(M,q_t) $}
 
 \For{$m\ \in\ M'$} 
 
 /* Compute prediction error estimate /*
 \State {$me\ \leftarrow\ m.generalization\_error$} 
 \State {$ex \gets m.unitary\_cost$}
 \State {$mf \gets m.framesize$}
 \State {$qtf \gets q_t.framesize$}
 \State {$c \gets cost\_function(me,mf,qtf, ex,\mu_e) $}
 \State {$ pq_i.push(c,<m,q_t>)$}
 \EndFor
\EndFor

/* Collect the best allocation */
\State {$S \gets \bigcup_{i=1}^{|query_T|} pq_i.top()$} 
\State {$Return \ S$} 
\EndFunction 
\end{algorithmic} 
\end{algorithm}

The constraints in Eq. \ref{eq:solutionConstraint} prune the search space significantly. We adopt a greedy algorithm that selects the candidate model that minimizes the cost function in Eq. \ref{eq:cost} for every tile. In line 3, we fork a thread that initializes a \textit{min priority queue} data structure for every tile $q_t$. This priority queue is sorted by the estimated cost, keeping the minimum cost at each instance as the top element in the queue. In line 6, we estimate the generalization error for every model on the current tile. Some models may have values for the prediction error or for the prediction time much farther from those in the set of available models. We consider these values to be outliers and drop the corresponding models from the set of candidates -- line 7 -- so that a normalization procedure can be applied to the values of the error predictions and prediction execution time in the cost formula. In line 8, we iterate through the set of candidate models, $m \in M'$. We evaluate the \textit{cost function} over the following arguments: the estimate for the model unitary execution time, the estimate for the generalization error, the frame size, and the weighting factor. The returned cost estimate is inserted into the priority queue in line 14. in line 17, we compose the optimal plan with the minimum allocation for every tile.

\section{Experiments}

In this section, we evaluate the assumptions considered in this work and the applicability of the DJEnsemble approach. The following questions are evaluated:
\begin{itemize}[leftmargin=*,noitemsep,nolistsep]
 \item Is the \textit{error\textunderscore function} satisfactory to estimate the generalization error of a predictor in a query region?
 \item Does the offline phase improve the estimate of the predictors' error under reasonable cost?
 \item Is the DJEnsemble approach resilient to variations in hyperparametrization and training data?
 \item How does the DJEnsemble approach behave against other ensembling approaches?
 \item Does the cost function enable the selection of accurate and efficient ensembles? 
\end{itemize}

\subsection{Setup}

We introduce the experimental scenario and methodology, the computational environment, and the implementation of DJEnsemble in SAVIME \cite{lustosa:2019}.

\subsubsection{Experimental scenario}\label{sec:methodology}

The data used in the experiments is a subset of the \href{http://www.data.nodc.noaa.gov}{Climate Forecast System Reanalysis (CFSR)} dataset that contains air temperature observations from January 1979 to December 2015 covering the space between 8N-54S latitude and 80W-25W longitude (temperature dataset) \cite{cfsr}. CFSR provides a homogeneous grid with four daily temperature observations. We concatenate these readings into a 3D grid with the structure ($day_i$, latitude, longitude, temperature). This grid can be interpreted as a continuous series of temperature behavior in the last 30 years. Additionally, we use a subset of the rainfall dataset from NASA's TRMM and GPM missions, with rainfall collected over 22 years (rainfall dataset) \cite{TRMM_GPM}. While the structure of this dataset is similar to the temperature dataset, there is a single daily observation. We select the same spatial regions in both cases.

\begin{figure}[!ht]
	\centering
	\includegraphics[width=0.33\textwidth]{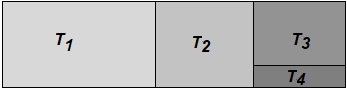} 
	\caption{Synthetic dataset. Each tile presents a slight variation in the data distribution.}
	\label{fig:query}
\end{figure}

We also build a synthetic dataset with a controlled data distribution variation. A tile $t_{i}$ is selected from the spatio-temporal data domain of the CFSR temperature dataset. The data distribution in $t_i$ is represented by its centroid time-series. Then, we partition it into four disjoint sub-regions $T_1(size = [10 \times 40]), T_2(size = [10 \times 30]), T_3(size = [8 \times 30])$, and $T_4 (size = [2 \times 30])$ -- as depicted in Figure \ref{fig:query} -- and 200 time slots. The size of each tile is given in terms of latitude and longitude. Every point within a tile is separated from the others by $0.5$ degrees and is indexed by a number. A region $[10 \times 10 \times 200]$ is a matrix covering 10 degrees of latitude and longitude, and having $200$ temporal temperature measurements at every spatial point. We add Gaussian noise to every sub-region in order to simulate variation in the data distribution. In Figure \ref{fig:query}, the gradation of the gray color shows the intensity of the added noise. The intensity varies from $0.1$ to $0.75$. The region covering tiles $t_i$, $1 \le i \le 4$, is the target of the predictive query $Q$. 

Thirty-six models are created using ConvLSTM architecture. Twenty-one of them are trained in the temperature domain. Models $SA_1$ to $SA_6$ share the same architecture -- filters, layers, etc. -- and are trained in different regions. A second set consisting of models $DA_1$ to $DA_7$ are trained in different regions and with different architectures. Model $SR_1$ is considered a baseline model and is trained in the region where the predictions to answer the predictive query are computed. The last seven temperature models and the fifteen models trained on the rainfall dataset are used to answer the five queries on real data in Section \ref{sec:queries_on_real_data}. All of them have different architectures.


\subsubsection{Computational environment}

The computing environment is kept constant throughout the experiments. It consists of a Dell PowerEdge R730 server with 2 Intel Xeon E5-2690 v3 @ 2.60GHz CPUs, 768GB of RAM, and running Linux CentOS 7.7.1908 kernel version 3.10.0-1062.4.3.e17.x86\_64. The models are trained and tested on an NVIDIA Pascal P100 GPU with 16GB RAM.

\subsubsection{SAVIME}
\label{sec:savime}

We integrate DJEnsemble into SAVIME, an in-memory columnar multidimensional array data management system \cite{lustosa:2019}. Given its multidimensional array data model, SAVIME is well suited for storing the spatio-temporal datasets used in the experiments. We have also implemented the complete weather prediction scenario presented in our motivating example from Section \ref{sec:Introduction} in SAVIME.

\subsubsection{Methodology}

Our evaluation of DJEnsemble considers the following metrics: accuracy, root mean square error, and composition execution time---as defined in Eq. \ref{eq:cost}. The model elapsed time ($ET$) is computed by averaging its prediction time for ten executions, at a single instance, for a unitary cost. The execution time ($E_xT$) is computed as $NE \times ET \times NTS$, where NE is the number of executions and $NTS$ is the time interval to forecast. The contribution of $E_xT$ in Eq. \ref{eq:cost} becomes relevant when considering a set of candidate models with different architectures (i.e., variations in the ConvLSTM network hyper-parameters).

The model accuracy term composing the cost function in Eq. \ref{eq:cost} is measured as the root mean square error (RMSE) of the predictions. The estimate (S) for a model allocation RMSE is obtained by executing the \textit{error\_function} for every allocation. Additionally, the real RMSE (R) is obtained by executing a model according to its allocation and comparing its predictions in a frame against the real values at every position of the query tile the frame has been allocated to and then applying the RMSE equation. The RMSE and execution time values are normalized by dividing them by the maximum observed value.

The evaluation of the DJEmsemble approach highlights three results. The estimated cost (S) is given by the cost function and is used to plan the execution. The real cost of the selected plan (R) quantifies the actual cost of running the model according to the allocation plan. The best execution (BE) gives the optimal allocation for the query.

\subsection{Results}\label{sec:results}

We present extensive experimental results to answer all the questions identified at the beginning of this section.

\begin{figure}[!ht]
 \centering
 \includegraphics[width=0.4\textwidth]{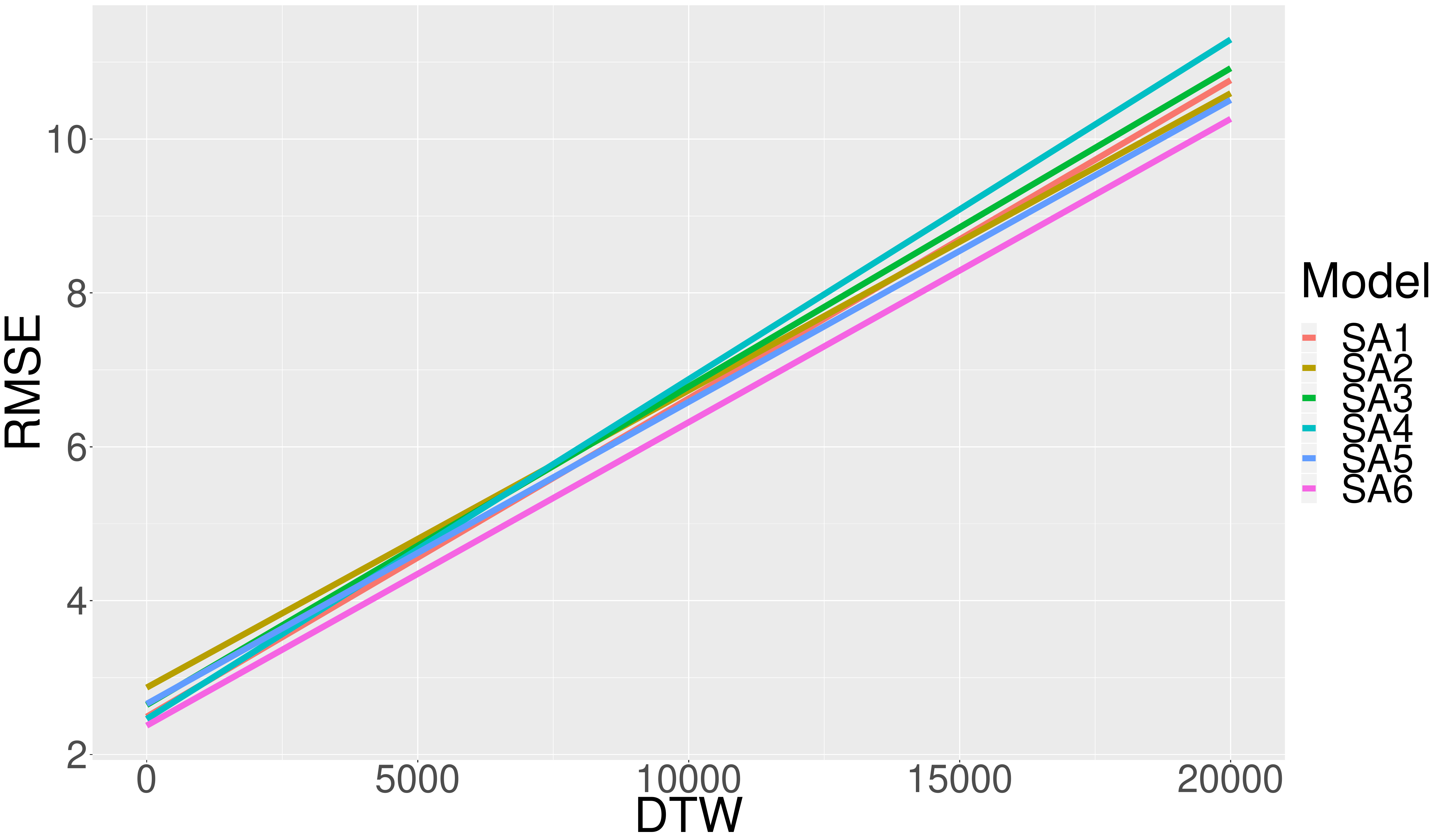}
 \caption{Generalization error on increasingly distant datasets for predictors with identical hyper-parametrization.}
 \label{fig:Fits1}
\end{figure}

\subsubsection{Distance to error model fitting}\label{sec:expmodelfitting}

\begin{figure}[!ht]
 \centering
 \includegraphics[width=0.4\textwidth]{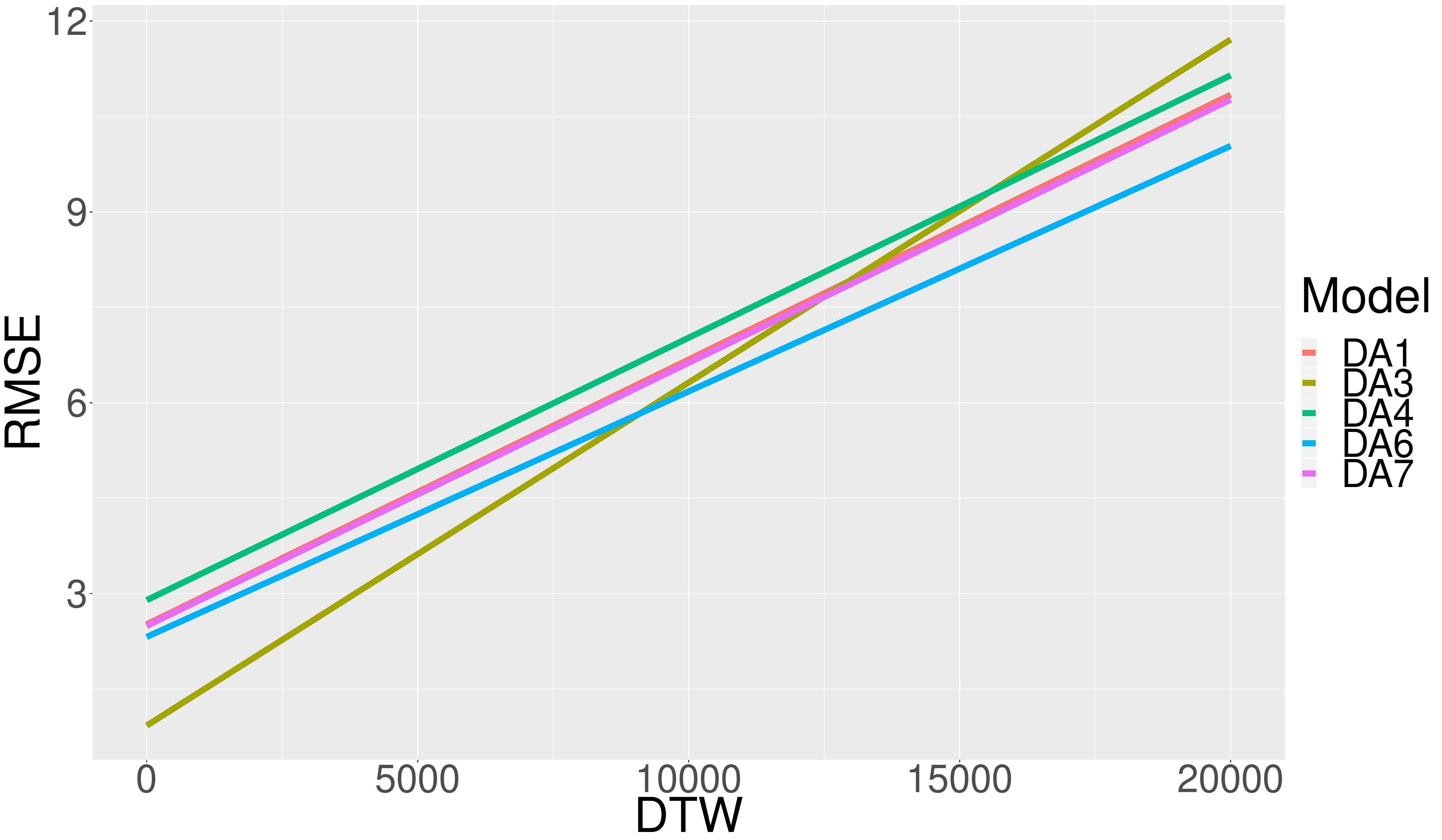}
 \caption{Generalization error on increasingly distant datasets for predictors with different hyper-parametrization.}
 \label{fig:Fits}
\end{figure}

An important assumption in this work is that we can predict the error of a black-box model on unseen data by a learning error function. In this experiment, we evaluate the \textit{error\-function} predictor. Figure \ref{fig:Fits1} and \ref{fig:Fits} depict the generalization error curve for a set of black-box STP models $M=\{DA_1,\ldots,DA_7, SA_1,\ldots,SA_6\}$ obtained by applying the procedure presented in Section \ref{sec:error-function}, on 50 datasets for each model. On the $y$ axis, we plot the estimate for the generalization error, considering the RMSE in the spatio-temporal region $d_i$ being predicted. The distance between the region $d_i$ and the base region $d_0$ -- computed using the \textit{DTW} function -- is the measure on the $x$ axis. We can observe that -- when the \textit{DTW} distance crosses the $10,000$ mark -- the RMSE accuracy clearly distinguishes the generalization capacity of the considered models. In Figure \ref{fig:Fits}, we show that the error computed by the \textit{error\_function} can be used to rank the models for a given STP prediction. Moreover, the results show that the error function reflects the generalization capacity of every model, as well as showing a strong correlation between distance and error---as the curves in both Figure \ref{fig:Fits1} and \ref{fig:Fits} are monotonically increasing.



\subsubsection{Feature-based vs shape-based clustering}
There are several alternatives for time-series clustering inroduced in previous work \cite{Aghabozorgi:2015}. As discussed in Section \ref{sec:clustering}, DJEnsemble adopts a feature-based approach, in which the GLD function is applied to the time-series as a feature extractor, followed by an Euclidean distance function computation between pairs of points. An alternative is to adopt a shape-based approach, in which the DTW function computes the distance between time-series. In both cases, once the distance between all the pairs of spatial points is computed, a clustering algorithm, such as k-means, can be applied. In this experiment, we compute the cost of building the distance matrix in both approaches over the rainfall dataset. Eq. \ref{eq:GLDDTWCOST} generalizes the correlation between the two costs, where $n$ is the number of time-series to be clustered:
\begin{equation}
  T_{shape}=68.4(n-1)T_{feature}
 \label{eq:GLDDTWCOST}
\end{equation}

To derive Eq. \ref{eq:GLDDTWCOST}, we perform an experiment where we randomly sample $n=\{100,200,300,400,500\}$ time series and build the corresponding distance matrix both by extracting features and computing the Euclidean distance between them, as well as by performing $n$ DTW calculations between the series and storing the elapsed time. Then, by fitting the elapsed time as a function of $n$ for both approaches and considering that to build the distance matrix of $n$ time series we need $\frac{n(n-1)}{2}$ calculations, we approximate the relationship derived in Eq. \ref{eq:GLDDTWCOST}.

\subsubsection{Effect of tiling}\label{Theeffectoftiling}

\begin{figure}[htbp]
 \centering
 \includegraphics[width=0.33\textwidth]{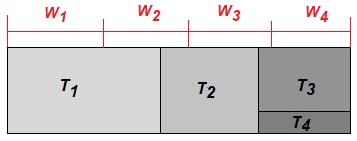}
 \caption{Aligned tiling $T_i$ and regular griding $W_i$.}
 \label{fig:Regular}
\end{figure}

As part of the DJEnsemble approach to solve the OSTEMPQ problem, we partition the data domain into aligned non-regular tiles. The tiling process partitions the data domain into hyper-rectangles of varying sizes. Every hyper-rectangle represents a spatio-temporal convex region, sharing a time-series with close data distribution among themselves. In this experiment, we conduct an ablation test, where we compare the DJEnsemble approach when applied to a region with and without data distribution based tiling. The latter considers a domain discretization using a regular grid. Thus, we fix the requirement of having the domain partitioned into convex regions while we relax the constraint of having cells sharing a close data distribution. In the following experiment, we consider a small variation from the spatio-temporal predictive query depicted in Figure \ref{fig:query}, as shown in Figure \ref{fig:Regular}. The cells marked as $T_i, 1 \leq i \leq 4$, correspond to the aligned tiling of the region, whereas the intervals denoted by $W_j, 1 \leq j \leq 4$, correspond to the tiling of the region as a regular grid. 

Table \ref{tab:tilingeffect} summarizes the prediction results of six different models -- trained in regions disjoint from the query region -- on every cell in Figure \ref{fig:Regular}. The values in Table \ref{tab:tilingeffect} reflect the RMSE on temperature values. When considering the plan corresponding to every domain partition, the best plan for the regular grid is:
\begin{equation*}
Best_W = [(W_1,DA_3),(W_2,DA_3),(W_3,DA_3),(W_4,DA_4)]
\end{equation*}
while for the aligned non-regular tiling, the best plan is:
\begin{equation*}
Best_T = [(T_1,DA_3),(T_2,DA_3),(T_3,DA_4),(T_4,DA_3)]
\end{equation*}
Although the best plans in the two solutions share similarities, the overall RMSE difference is $14.1\%$ in favor of the aligned non-regular tiling. This is reassuring as the more precise domain partitioning leads to better spatial model positioning---resulting in improved prediction accuracy.

 
\begin{table}[htbp]
	\begin{center}
	\begin{small}
	\setlength{\tabcolsep}{2.5pt}
		\begin{tabular}{@{}lrrrrrrrrrrrr@{}}\toprule
 M&$W_1$ & $T_1$ & $W_2$ & $T_2$ & $W_3$ & $T_3$ & $W_4$ & $T_4$ \\ \midrule
 $DA_1$&19.33 & 20.23 & 22.21 & 23.57 & 24.02 & 24.68 & 24.54 & 23.79 \\ \hdashline
 $DA_2$&81.07 & 81.09 & 80.84 & 80.73 & 80.72 & 80.7 & 80.71 & 80.74 \\ \hdashline
 $DA_3$&\textcolor{red}{4.36} & \textcolor{red}{3.61} & \textcolor{red}{2.14} & \textcolor{red}{1.69} & \textcolor{red}{1.92} & 2.43 & 2.58 & \textcolor{red}{2.21} \\ \hdashline
 $DA_4$&6.38 & 5.17 & 3.79 & 2.79 & 2.46 & \textcolor{red}{2.11} & \textcolor{red}{2.56} & 2.91 \\ \hdashline
 $DA_5$&81.35 & 81.3 & 81.17 & 81.1 & 81.1 & 81.05 & 81.09 & 81.11 \\ \hdashline
 $DA_6$&19.08 & 20.1 & 21.98 & 23.49 & 23.88 & 24.53 & 24.51 & 23.75 \\
			\bottomrule
		\end{tabular}
	\end{small}
\caption{Comparison between aligned tiling and regular griding in terms of the RMSE error.}
\label{tab:tilingeffect}
	\end{center}
\end{table}


\subsubsection{Effect of STP characteristics}\label{sec:varyingarchitecture}

The next set of experiments is divided into three groups, where we increasingly vary the difference among the black-box STPs. We consider the deep net hyper-parametrization, the training data distribution, and the query region data distribution.

\paragraph{Single architecture, varying training data distribution, and multiple tiles for prediction}

In this experiment, we consider a query over multiple tiles, for which we assess the combination of models that minimizes a cost function composed by the prediction error and the model execution cost. We fix the architecture for the model set $M=\{SA_1, SA_2, \ldots, SA_6\}$. These models are trained on different domain regions, which do not match the query region. As the models are similar in terms of architecture and input data frame size, the execution elapsed time, and the number of model instance invocations required to answer the query are constant per tile for all models. The DTW distance is used to inform on the distance between the model training region and the query regions---used as input in estimating the generalization error on a query tile.




\begin{table}[htbp]
\centering
	\begin{small}
		\begin{tabular}{@{}lrrrrrrrrrrrr@{}}\toprule
 \textbf{$Model$} & \textbf{$T_{1}$} & \textbf{$T_{2}$} & \textbf{$T_{3}$}& \textbf{$T_{4}$}\\
 \midrule
 $SA_{1}$&2008.88 & 2640.48 & 3215.50 & 2150.72\\\hdashline
 $SA_{2}$&49763.79 & 57445.83 & 61182.64 & 55608.07\\\hdashline
 $SA_{3}$&1813.04 & 2737.75 & 3200.94 & 2096.85\\\hdashline
 $SA_{4}$&1849.48 & 2519.36 & 3465.86 & 2220.04\\\hdashline
 $SA_{5}$&49554.81 & 57122.75 & 60786.23 & 56020.07\\\hdashline
 $SA_{6}$&47442.06 & 55341.19 & 57588.49 & 54560.79\\\hdashline
			\bottomrule
		\end{tabular}
	\end{small}
\caption{DTW distance between the data tiles used in the training of models $SA_1$ to $SA_6$ and the query tiles.}
\label{tab:models1}
\end{table}

First, we interpret the distances in Table \ref{tab:models1} with respect to the four query tiles. Table \ref{tab:models1} highlights that the black-box STPs $SA_1$, $SA_3$, and $SA_4$ are trained in the regions whose data distribution is close to that of the query region, leading to smaller distances. The training region for model $SA_3$ shows the smallest distance among all the models, except on $T_2$, where $SA_4$ has an even closer distribution. The complement to the distance information is given by the \textit{error\_function}. As shown before in Figure \ref{fig:Fits1}, the generalization capacity of $SA_1$ is constantly higher than the other models. Thus, a decision based on Figure \ref{fig:Fits1} picks model $SA_1$ for all the query tiles, whereas the distance-based information weighs toward $SA_3$ and $SA_4$.

\begin{figure}[htbp]
	\includegraphics[width=0.48\textwidth]{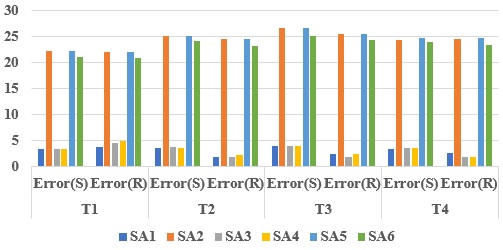}
	\caption{The prediction error is estimated by the cost function (S) and evaluated on the query tile (R). Same hyper-parameters and different training datasets.}
	\label{fig:models_error_sa}
\end{figure}

\begin{figure}[htbp]
	\includegraphics[width=0.45\textwidth]{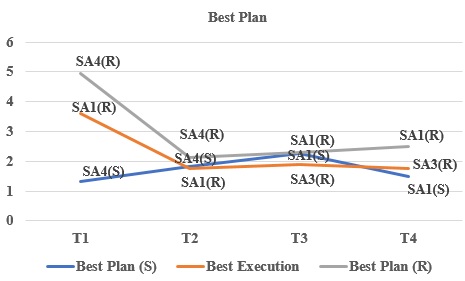}
	\caption{RMSE comparison between the best plans chosen by DJEnsemble and Best Execution.}
	\label{fig:models_plan_sa}
\end{figure}

Figure \ref{fig:models_error_sa} depicts both the cost model estimate (S) and the real error (R) for every model at every query tile. We observe that the cost model can easily discard models $SA_2$, $SA_5$, and $SA_6$. Additionally, the three remaining candidates are the ones with the best execution results. Thus, the cost model narrows down the set of candidate models to be used to answer the query. Figure \ref{fig:models_plan_sa} summarizes the results from a complete query region prediction point of view. The blue line corresponds to the plan chosen by the cost model. The gray line corresponds to the actual performance results obtained when running the cost function chosen plan. Finally, the orange curve depicts the actual best plan based on real errors. The separation between the gray and the orange line gives an idea of the real loss in quality that we incur when using the execution plan selected by our algorithm. As can be observed, the predictions are very close to the actual error---proving a close calibration of the cost model and the error function.

\paragraph{Multiple architectures, varying training data distribution, and multiple tiles for prediction}

We select seven models with different architectures, $DA_1$ to $DA_7$, and having different execution time and accuracy. Different from the previous experiment with models having the same architecture, in this scenario, the number of invocations in each tile is important as it varies from one model to another. The number of model invocations is associated with the spatio-temporal region covered by it and the corresponding region of the query area to be predicted.

\begin{figure}[htbp]
	\includegraphics[width=0.48\textwidth]{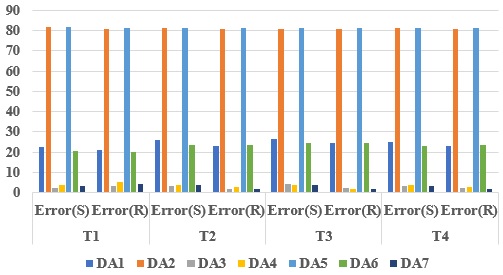}
	\caption{The prediction error is estimated by the cost function (S) and evaluated on the query tile (R). Different model architecture and training datasets.}
	\label{fig:models_error_da}
\end{figure}

\begin{figure}[htbp]
	\includegraphics[width=0.45\textwidth]{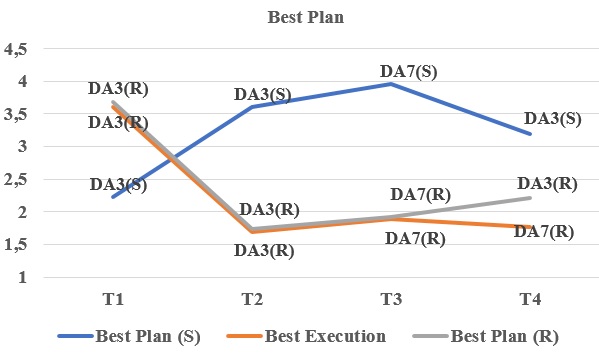}
	\caption{RMSE comparison between the best plans chosen by DJEnsemble and Best Execution.}
	\label{fig:models_plan_da}
\end{figure}

Figure \ref{fig:models_error_da} summarizes the estimated and actual errors of every model by the query tiles. The first observation is that the error estimate is again very close to the actual computed errors. The models with the closest data distribution -- $DA_7$ and $DA_4$ -- are the ones that indeed exhibit the best accuracy. Additionally, Figure \ref{fig:Fits} shows that $DA_3$ exhibits a very good generalization capacity up to a DTW distance of $9,000$. Thus, it also qualifies as a competitive model in our estimates. Figure \ref{fig:models_plan_da} shows the best plan, considering the cost function estimates (blue line), real execution of the cost-based chosen plan (gray line), and overall best execution (orange line). On tiles $T_1$, $T_2$, and $T_3$, our cost model choice matches the best actual execution---$(T_1,DA_3)$, $(T_2,DA_3)$, and $(T_3,DA_7)$. On $T_4$, the error of $DA_3$ and $DA_7$ are close to $1.76$ and $2.21$, respectively. We must remember that tiles $T_3$ and $T_4$ with an area $(8\times 30)$ and $(2\times 30)$, respectively, are smaller than the input size of the $DA_3$ ($10\times 20$) and $DA_7$ ($10\times 10$) models. In this case, it is necessary to extend the borders of the tile on the latitude dimension and this causes the distortion of the real error. 

We compare our allocation:
\begin{equation*}
[(DA_{3},T_1), (DA_{3},T_2), (DA_{7},T_3), (DA_{3},T_4)] = 2.35
\end{equation*}
with the best allocation that can be generated from the set of available models:
\begin{equation*}
[(DA_{3},T_1), (DA_{3},T_2), (DA_{7},T_3), (DA_{7},T_4)] = 2.24
\end{equation*}
In this case, DJEnsemble obtains a plan that is $4.91\%$ less accurate than the best model composition. Thus, we observe that models with varying generalization error and hyper-parameters can also be combined through the DJEnsemble approach. Moreover, the contribution of the execution time on the choice of the best plan is not significant, as the fastest models are also $DA_3$ and $DA_7$---which exhibits the best accuracy in this experiment.

\subsubsection{DJEnsemble versus baseline approaches}\label{sec:baseline}

We consider two baseline approaches to solve the OSTEMPQ problem. The first approach takes a single model trained on the same region as the query and uses it as a predictor for the query (i.e., single model baseline). The second approach applies the traditional ensemble technique (i.e., ensemble baseline), as presented in Section \ref{sec:composition-approach}.

\paragraph{Single model baseline}
The single model baseline is constructed as follows. We use the $SA_1$ model architecture -- which works best in most of the experiments -- and train, validate, and test it on the same region over which query $Q$ is specified using the time interval 1 to time-final (200). The remaining time slots are used for the query.

\begin{table}[htbp] 
\centering
	\begin{small}
		\begin{tabular}{@{}lrrrrrrrrrrrr@{}}\toprule
			\textbf{DJEnsemble Plans} & \textbf{Error} & \textbf{Improvement}\\ \midrule
			1- (`$DA_{3}$', `$DA_{3}$', `$DA_{7}$', `$DA_{3}$') & 2.35 & 18\%\\ \hdashline
			2- \underline{(`$DA_{3}$', `$DA_{3}$', `$DA_{7}$', `$DA_{7}$')} & 2.24 & 21.4\% \\\hdashline 
			 \textbf{Single model baseline/$SR_1$} & 2.85 & -\\
			\bottomrule
		\end{tabular}
	\end{small}
\caption{DJEnsemble vs single model baseline.}
\label{tab:models5}
\end{table}

Table \ref{tab:models5} summarizes the results of the best execution plan (DJEnsemble) against the single model baseline created on the query data distribution. The first plan (1) is detected by our cost function. The underlined plan (2) corresponds to the best performing plan when we take into account all the STP models. The values correspond to the accuracy error produced by executing the plans. This result confirms that DJEnsemble can find a composition of STP models exhibiting an accuracy improvement of $18\%$ when compared against a single model baseline that is built over the query region. The intuition for this result is that the disjoint ensemble of models offers a finer grain allocation of models to the query region. Conversely, the single model baseline approach learns different patterns existing in the training data of the query region, leading to less accurate predictions.

\paragraph{Ensemble baselines}
The second baseline includes multiple ensembles models. We build four ensembles and compare them against DJEnsemble and the single model baseline. The ensemble models are constructed from the seven models with different architectures $DA_1$ to $DA_7$. The traditional ensemble is built over all of them. The second ensemble selects three models considering a cutting threshold of $5$ degrees, computed using the error-function over DTW distances at every tile. However, the allocation does not obey the tiling of the query region. The third ensemble extends the second ensemble by using tiling as a guide to model allocation. Finally, the fourth ensemble creates a stacking from the seven base models. Since it is difficult to determine where the predictive query is positioned and which models are selected to solve it, a pre-trained ensemble does not work correctly. To tackle this issue, we create a runtime stacking ensemble with the candidate models to solve the query. We use previous data from the query region, generate the predictions of every model for that region prior to the query, and build a stacking by training a multiple linear regression model (MLR) that weighs every model correspondingly. Then, we generate the query prediction for every model and refine it using our pre-trained stacking. We build two DJEnsemble plans. The (S) plan is a composition selected according to the approach based on estimated errors. DJEnsemble (R) corresponds to a plan selected considering real errors.


\begin{table}[htbp] 
\label{tab:ensemble}
\centering
	\begin{small}
		\setlength{\tabcolsep}{1.5pt}
		\begin{tabular}{@{}lrrr@{}}\toprule
			\textbf{Ensemble approaches} & \textbf{Error} & \textbf{Perf.} & \textbf{Exec.Time}\\ \midrule
			1- Traditional ensemble & 21.03 & -838.83\% & 68.35 \num{+-0.586}\\ \hdashline
			2- Ensemble-DTW distance & 3.07 & -37.05\% & 34.38 \num{+-0.482}\\\hdashline 
			3- Ensemble-DTW and tiles & 2.68& -19.64\% & 43.46 \num{+-0.262}\\ \hdashline
			4- Stacking(MLR)-DTW distance & 2.92 & -23.28\% & 35.32 \num{+-0.301}\\\hdashline
			5- Single model baseline/$SR_1$& 2.85 & -21.00\% & 4.22 \num{+-0.059}\\
			 \hdashline
			6- DJEnsemble (S) & 2.35 & -4.91\% & 14.06 \num{+-0.193}\\ \hdashline 
			7- \textbf{DJEnsemble (R)} & 2.24 & -\\
			\bottomrule
		\end{tabular}
	\end{small}
\caption{DJEnsemble vs ensemble baselines.}
\label{tab:models9}
\end{table}
 
The results obtained by every ensemble we consider are included in Table \ref{tab:models9}. We observe that the more specific the allocation is, the more accurate the prediction becomes. Thus, capturing the data distribution in the domain tiles and using it as a guide for selecting and allocating models pays off. We also observe a considerable difference in latency. It is clear that the traditional ensemble approach requires running every selected model over the entire query region. Every model is invoked as many times as needed to cover the spatial region of the query. Additionally, the traditional ensemble requires a post-processing action to aggregate the values per prediction point and compute the average of the results. This leads to a performance penalty of almost $9X$. It is interesting to observe that the usage of DTW in filtering models in ensembles (2) and (4) significantly reduces the execution cost, as fewer models are run. Moreover, this makes the ensemble more specific, contributing to a more precise prediction. Moving incrementally toward DJEnsemble, ensemble (3) applies filtering and allocation per tile and shows competitive performance results. However, its latency is slightly inferior to the one observed for (2). This is due to a more frequent change in the execution context, as models are run per tile. The single model baseline (5) has the same spatial input as the size of the query. Finally, we present the DJEnsemble errors for a model allocation based on estimated errors with our cost function (6) and with an ideal allocation (7). From the perspective of computational resources, every invocation of the prediction function takes up less than 256 MB of memory. This allows any of the DJEnsemble plans to be executed in parallel at the same time as the prediction time of the slowest model.




\subsubsection{Cost model analysis}

In the following, we demonstrate how the user can specify a preference for more accurate predictions or lower execution time models by parameterizing the cost function. We run experiments with the queries in Section \ref{sec:queries_on_real_data} and define the performance metric $M_e$ in terms of both the estimated error and the estimated model prediction time. We consider the total number of times a model has to be invoked in order to generate predictions for a given tile, as shown in Eq. \ref{eq:cost}.


\begin{table}[htbp] 
\centering
	\begin{small}
	\setlength{\tabcolsep}{3.8pt}
		\begin{tabular}{@{}crrrrrr@{}}\toprule
			 \textbf{$\mu_e$} & \textbf{Query 1}& \textbf{Query 2} & \textbf{Query 3} & \textbf{Query 4} & \textbf{Query 5} \\ \midrule
			 0.0 & 3.35 & 4.29 & 5.34 & 6.04 & 3.18 \\\hdashline 
			 0.2 & 4.38 & 4.80 & 5.32 & 6.01 & 4.96 \\\hdashline 
			 0.4 & 4.67 & 6.19 & 5.14 & 5.98 & 5.76 \\\hdashline 
			 0.6 & 6.07 & 8.98 & 5.13 & 5.95 & 5.86 \\\hdashline 
			 0.8 & 7.14 & 10.26 & 5.12 & 5.91 & 7.86 \\\hdashline 
			 1.0 & 12.31 & 19.18 & 4.64 & 5.55 & 20.08 \\\hdashline 
			\bottomrule
		\end{tabular}
	\end{small}
\caption{RMSE sensitivity with respect to the cost function.}
\label{tab:rmse_time_varying_data}
\end{table}

The results of this experiment are included in Table \ref{tab:rmse_time_varying_data}. The results demonstrate that -- for queries run on both datasets -- the value of $\mu_e$ directly impacts the prediction error of the chosen models. Queries executed on the temperature dataset clearly tend to present a higher prediction error as the importance of the prediction time estimate increases. This is expected since, in this case, the DJEnsemble algorithm tends to choose models with higher estimated error in exchange for smaller execution time. However, for queries 3 and 4 executed on the rainfall dataset, the prediction error tends to become slightly smaller as the value of $\mu_e$ increases. This is due to the fact that, as the execution time becomes more relevant to the cost function, larger area models are chosen. These models tend to be invoked a smaller number of times when allocated to larger tiles, capturing spatial correlations that would not be possible otherwise. The different spatial resolution across the two datasets explains why the same effect is not observed for the other queries.


%

\subsubsection{DJEnsemble performance breakdown}

We investigate the performance cost associated with both phases -- offline and online -- of the DJEnsenble approach.

\paragraph{Offline cost}
In this section, we present the computational cost of executing the offline phase of DJEnsemble. The offline phase consists of two separate processes---registering a dataset and registering an STP model. The cost of these processes is modeled as:


\begin{equation}
Cost_{Dataset}= GLD + Clustering + Tiling
 \label{eq:OfflineCostDataset}
\end{equation}
\useshortskip
\begin{equation}
Cost_{Model}= ApplyNoise + ApplyModel + Fit
 \label{eq:OfflineCostModel}
\end{equation}

We compute $Cost_{Dataset}$ for the preprocessing of the rainfall dataset composed of $200\cdot 200\cdot 21$ time-series, each consisting of 365 timestamps, with a total size of 2.5 GB. We choose a seasonality of one year because we are working with meteorological variables \cite{Nwogu2016,Iwueze2011}. In this case, $Cost_{Dataset}$ is approximately $84$ minutes. $Cost_{Model}$ computes the elapsed time for registering a model, considering its evaluation on 50 different datasets when building the learning function. On the rainfall dataset, $Cost_{Model}$ is approximately 25 seconds. While the total offline cost approaches $85$ minutes, it is important to emphasize that this is a one-time cost---dominated by the dataset preprocessing cost. Such a cost is common for the analysis of any new dataset.

\paragraph{Online cost}
The DJEnsemble online model selection and allocation are depicted in Algorithm \ref{alg:online}. The cost of this online processing can be expressed as:
\begin{equation}
 Cost_{query} = (DTW + LF) \cdot Nmodels \cdot NTiles
 \label{eq:OnlineCost}
\end{equation}
where $DTW$ is the cost of running a DTW distance between a pair of time-series and $LF$ is the cost of running the polynomial equation associated with the model. $Nmodels$ is the number of candidate models and $Ntiles$ is the number of tiles contained in the query region. For example, query $Q3$ over the rainfall domain considers 175 tiles and 14 models. It has an online cost $Cost_{query}$ of 735 seconds. Query $Q4$, which considers 94 tiles and the same 14 models, has an execution cost of only 384 seconds. These numbers represent $13\%$ ($Q3$) and $7.3\%$ ($Q4$) of the offline preprocessing cost.



\begin{figure}[htbp]
 \centering
	\includegraphics[width=0.48 \textwidth]{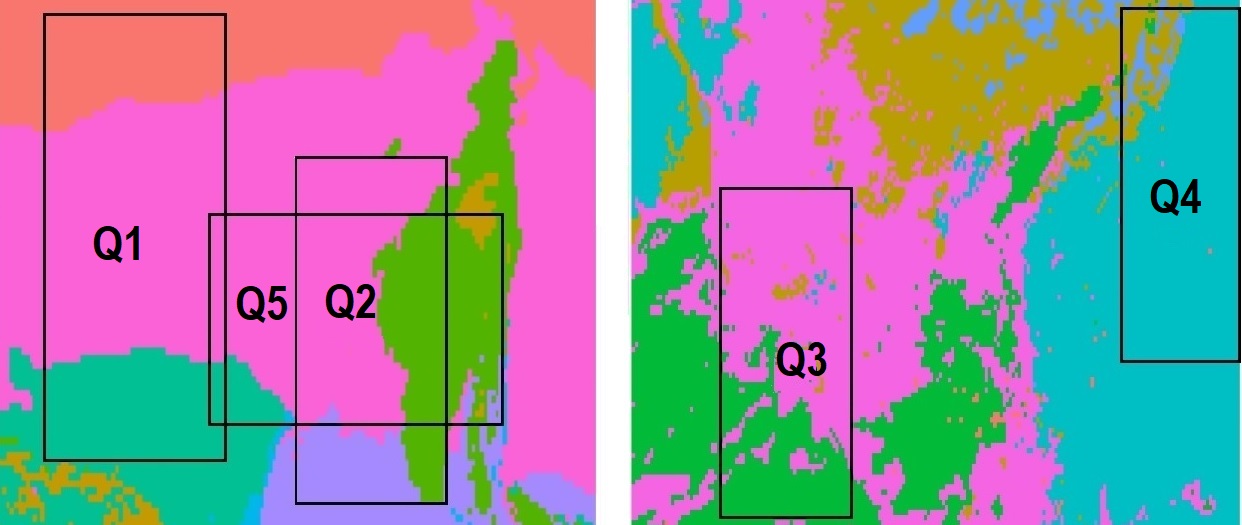}
	\caption{Queries over the temperature (left) and rainfall (right) domains.}
	\label{fig:real_queries}
\end{figure}

\subsubsection{Queries on temperature and rainfall datasets}
\label{sec:queries_on_real_data}

We present the results obtained on five queries executed on real data extracted from the temperature and rainfall domains. Temperature is considered by meteorologists as an easy variable to model as opposed to rain, which is considered one of the most complex variables to predict. The temperature is homogeneously distributed in large spatial regions, while the rainfall is more heterogeneous and concentrated in small areas. Based on these, we can argue that we define queries on data with different behavior. Figure \ref{fig:real_queries} depicts the regions corresponding to every query. Each color represents a region with different data distribution. Table \ref{tab:rmse_real_data} summarizes the results for a traditional ensemble, a stacking ensemble, and DJEnsemble. DJEnsemble achieves the best accuracy for all the queries. The gap between DJEnsemble and the other ensembles is as much as a factor of $9$ or more.

\begin{table}[htbp] 
\centering
 \resizebox{\columnwidth}{!}{
		\begin{tabular}{@{}lrrrrrr@{}}\toprule
			\textbf{Query} & \textbf{R[lat,lon]}& \textbf{Trad. ensemble} & \textbf{Stacking} & \textbf{DJEnsemble} \\ \midrule
			 Q1 & [70:130,95:140] & 25.01 & 7.28 & 3.35 \\\hdashline 
			 Q2 & [60:110,40:80] & 27.88 & 5.52 & 4.29\\\hdashline
			 Q3 & [125:175,25:90] & 14.28 & 13.96 & 5.34\\ \hdashline 
			 Q4 & [0:40,60:130] & 8.80 & 12.94 & 6.04\\ \hdashline 
			 Q5 & [59:100,25:100] & 29.10 & 5.04 & 3.18\\ \hdashline 
			\bottomrule
		\end{tabular}
	}
\caption{RMSE over the temperature and rainfall domains.}
\label{tab:rmse_real_data}
\end{table}


\section{Related Work}

DJEnsemble is an automatic approach for the selection and allocation of models to compose an ensemble for spatio-temporal predictions. This approach is orthogonal to the design and exploration of specific learning strategies and algorithms. It can be applied to a set of spatial models implementing algorithms varying from time-series predictors -- such as ARIMA -- to spatio-temporal deep learning models with different architectures and hyperparameters \cite{shi2015convolutional,Cheng:2018,Yania:2018}. Additionally, DJEnsemble has been conceived for the prediction of auto-regressive problems. For more complex spatio-temporal forecasts, such as urban traffic for points of interest detection, road networks, and people mobility, more sophisticated deep learning solutions \cite{Pan:2019} are required.

\subsection{Meta-Feature Learning}

In DJEnsemble, model selection follows a meta-learning approach \cite{brazdil:2009,Prudencio:2004,Vilalta:2001}---datasets are used in learning a model that predicts deep learning models' performance, enabling model selection. This is similar to the approach developed in AutoGRD \cite{Cohen-Shapira2019}, which is designed for independent multi-variate based prediction. AutoGRD jointly considers an embedding of dataset instances and information about the performance of models on these datasets to learn a model performance predictor. The latter is used to predict the models' performance on unseen datasets. DJEnsemble deals with a similar challenge in learning spatial-temporal patterns and inferring models' performance in any region of the spatial domain. In building the ensemble, DJEnsembles faces an additional challenge in devising a strategy for the spatial allocation of the models. We adopt the tiling strategy on clustered time-series in order to address this challenge. Regarding the pre-processing step for model selection, AutoGRD uses an ensemble of trees to infer dataset instances co-occurrence and distances between instances. This is analogous to the offline phase of DJEnsemble, which informs on the distances between domain regions. However, although adopting a similar meta-learning approach for model selection, AutoGRD selects a single best performing model. As we have shown in Section \ref{sec:baseline}, DJEnsemble outperforms a single model trained in the same region as the query by 18\%, due to its ability to decide on spatial model allocation through the tiling of the clustered domain.


\subsection{Ensembles}

There are several approaches that apply deep learning techniques as model stacking strategies to improve the performance of base models \cite{Ambrogioni2017,Xiao2017,Yania:2018}. Other ensemble approaches use a single deep learning architecture. In this type of strategy, while some layers of the architecture learn, others are used for second-level supervision acting as an ensemble \cite{Hu2017,Han2016}. Other strategies are ensemble-based black-box attacks to explore the vulnerability of the deep learning models---which is significant to choose effective substitute models for ensembles. The name black-box comes from the premise that the model's architecture is not known. The decision boundaries of the primary model are explored through constructed examples and -- for cases outside this data space -- substitute models are trained \cite{HANG2020}. The main difference between the approach we propose in the online phase and the existing literature is that our decisions are data-driven (i.e., data distributions) and our objective is to solve an auto-regressive problem.

\subsection{AutoML and Model Serving Systems}

DJEnsemble contributes to the broader topic of AutoML \cite{AutoML:2019} with respect to model selection. This approach has been introduced in various state-of-the-art predictor serving systems \cite{Kang2017,noscope++,deluceva}. The framework Clipper \cite{Crankshaw2017} is designed to serve trained models at interactive latency. It implements two model selection policies based on multi-armed bandit algorithms. Both policies span a trade-off between accuracy and computation overhead with adaptable batch sizes. Rafiki \cite{Wang2018} is a machine learning training and inference service. It provides an online multi-model selection to compose ensembles for multiple requests. Rafiki uses a reinforcement learning approach to reward accuracy and penalize overdue requests. In this sense, DJEnsemble considers a cost model that statically defines a prediction ensemble plan based on estimates for accuracy and prediction time. In fact, DJEnsemble can be implemented as a model selection solution if these systems provide a service for auto-regressive STP predictions.




\section{Conclusions and Future Work}

This paper presents DJEnsemble, a disjoint ensemble approach to plan for the composition of black-box deep learning models to answer spatio-temporal auto-regressive predictive queries. DJEnseble includes an offline phase -- where data are partitioned into tiles and a learning function is computed for every STP -- and an online phase---where a cost function is applied to rank candidate models to be allocated to query tiles, considering an estimate for the generalization error and the model inference time. The experimental results show that DJEnsemble significantly outperforms traditional ensemble strategies in both accuracy ($9X$) and execution time ($4X$). Overall, DJEnsemble produces more accurate predictions than all the other alternatives. 

There is plenty of future work to be explored. The offline phase can be further optimized, improving the identification of spatio-temporal patterns and reducing the pre-processing cost. Techniques to improve the learning function accuracy -- especially in higher dimensions -- can also be investigated. The execution of the selected plans can take advantage of parallelism in the AI inference framework. Improvements in model design can also contribute to the overall prediction quality, especially considering the effect of different grid scales across datasets. Finally, we also plan to explore multivariate predictions and how to adapt the data distribution distance-based approach to this scenario.


\section{Acknowledgments}
The authors would like to thank CAPES, FAPERJ, and CNPq for scholarships and research productivity fellowships. We also thank Petrobras, Ger\^encia de Perfura\c c\~ao e Completa\c c\~ao de Po\c cos, and ANP for financing this work through contract 5850.0108 913.18.9. Finally, Fabio Porto thanks the National University of Singapore and Prof. Beng Chin Ooi for hosting him during the paper preparation.


\balance

\bibliographystyle{abbrv}
\bibliography{vldb_sample}

\end{document}